\documentclass[preprint,12pt]{elsarticle}

\usepackage{amssymb}

\usepackage{booktabs}
\usepackage{pdflscape}
\usepackage{algorithm}
\usepackage[noend]{algpseudocode}
\usepackage{url}
\usepackage{caption}
\usepackage{subcaption}
\usepackage{color}

\journal{Future Generation Computer Systems}

\begin{document}

\begin{frontmatter}



\title{Evaluation and Efficiency Comparison of Evolutionary Algorithms for Service Placement Optimization in Fog Architectures}


\author{Carlos Guerrero\corref{mycorrespondingauthor}}
\ead{carlos.guerrero@uib.es}
\cortext[mycorrespondingauthor]{Corresponding author}

\author{Isaac Lera\corref{}}
\ead{isaac.lera@uib.es}

\author{Carlos Juiz\corref{}}
\ead{cjuiz@uib.es}

\address{Crta. Valldemossa km 7.5, Palma, E07121, SPAIN}

\address[mymainaddress]{Computer Science Department, University of Balearic Islands}

\begin{abstract}
This study compares three evolutionary algorithms for the problem of fog service placement: weighted sum genetic algorithm (WSGA), non-dominated sorting genetic algorithm II (NSGA-II), and multiobjective evolutionary algorithm based on decomposition (MOEA/D). A model for the problem domain (fog architecture and fog applications) and for the optimization (objective functions and solutions) is presented. Our main concerns are related to optimize the network latency, the service spread and the use of the resources. The algorithms are evaluated with a random Barabasi-Albert network topology with 100 devices and with two experiment sizes of 100 and 200 application services. The results showed that NSGA-II obtained the highest optimizations of the objectives and the highest diversity of the solution space. On the contrary, MOEA/D was better to reduce the execution times. The WSGA algorithm did not show any benefit with regard to the other two algorithms.
\end{abstract}


\end{frontmatter}



\section{Introduction}
\label{intro}
In the last few years, there has been a significant increase in the number of applications developed for the Internet of Things (IoT). In these ecosystems, any device, however small, is able to connect to the Internet and to monitor or control physical elements. These devices gather, process, and share data. As these environments have become popular, their needs for processing and storing data have increased rapidly, making it impossible to use the devices themselves to meet all storage and processing requirements.

A first solution to provide higher capacities to the IoT applications was to integrate cloud services by storing and processing the data of the IoT devices in the cloud providers and sending the results back to the IoT devices. This earlier solution allowed to develop applications that interact with real devices, thanks to the IoT paradigm, and which have unlimited storage and processing capabilities, thanks to the cloud. But important drawbacks emerged related to the high communication delays and the use of the network to send high quantities of data between the IoT devices and the cloud providers. 

Meanwhile, network communication devices have increased their computational capacities, and a new paradigm has emerged, the fog computing~\cite{Mahmud2018}. A fog architecture provides computational and storage capacities in the intermediate in-network devices and in continuity with the cloud~\cite{openfog}. By this, the communication nodes are able to allocate services or to store data. This reduces the network latency of the applications since the services can be placed closer to the IoT devices. The data transmitted through the network is also reduced since it is processed and stored closer to the IoT devices. 

A similar paradigm is the edge computing, where the nodes of the edge of the network are the only ones with the capacity to execute the services. Sometimes, some services are only allocated in the cloud due to, for example, resource constraints, and the edge devices only allocate a subset of these services. Contrary to the fog computing, this second paradigm is more focused towards the \textit{things} side, instead of the \textit{cloud} and \textit{infrastructure} side~\cite{7488250}. This is characterized by using devices with more limited resources and making data sharing more challenging.

Important challenges and open research problems have emerged with the definition of fog architectures~\cite{8100873}. In particular, the placement of services and data gets an important role in the optimization of the systems~\cite{Velasquez2018}. Data and service management policies are needed to decide when and which fog devices the services should be allocated in. The problem of the selection of the fog devices where the services are instantiated is usually named as fog service placement problem (FSPP). Several service placement studies have established the first steps of future optimal solutions but there is still room for improvement.

Evolutionary optimization is a paradigm inspired by natural evolution mechanisms. The use of evolutionary algorithms is an emerging trend for effective and efficient optimization of complex systems~\cite{zhan2015cloud}. An important number of studies have addressed resource management problems with evolutionary algorithms and they are commonly studied for the increasing structural complexity of distributed architectures such as cloud or fog computing~\cite{guzek2015survey}. For example, Zhan et al. stated in their survey that around 18\% of the studies about cloud and resource management are concerning the use of evolutionary methods. Evolutionary algorithms have resulted in suitable solutions for the optimization of resource management in VM allocation~\cite{6701467}, data replica placement~\cite{Guerrero2018}, federated clouds~\cite{kimovski2016multi}, or software composition~\cite{frey2013search}, among others.

We study the suitability of three evolutionary algorithms for the problem of fog service placement: single-objective genetic algorithm with weighted sum transformation (WSGA); non-dominated sorting genetic algorithm-II (NSGA-II); and, multi-objective evolutionary algorithm based on decomposition (MOEA/D). Our study concerns the adaptation of the three algorithms to our specific domain, by defining the evolutionary operators and the parametrization of the algorithms. This is not the first study that compares the performance of those algorithms~\cite{5346628,Peng2009,4633340} but, to the best of our knowledge, it is the first one that applies them to the specific domain of fog architectures.

The rest of the paper is organized as follows: Section~\ref{relatedsect} reviews previous work related to fog service placement; Section~\ref{problemstatement} describes our model for the domains and our optimization objectives; Section~\ref{evolutionary} details the evolutionary approaches we have included in our paper; Section~\ref{sect:expvalidation} explains the details of the experimental study we performed; Section~\ref{resultdiscussion} presents the results and analyzes them; and, finally, Section~\ref{conclusion} presents the conclusions and possible future works.

\section{Related work}
\label{relatedsect}

The use of evolutionary approaches is widely adopted for resource management in distributed architectures~\cite{zhan2015cloud,guzek2015survey}. But their applicability in fog service placement problems is not enough researched as we explain in the following paragraphs. 

Some frameworks have been already implemented to facilitate the adoption of fog-based solutions in IoT applications. For example, the frameworks in Donasollo et al.~\cite{donassolo:hal-01859695} or in Skarlat et al.~\cite{8603163}. These frameworks incorporates simple policies for the orchestration and operation of the fog applications, but they enable the integration of new and more sophisticated policies in real fog domains. Challenges in the field of the resource management are now the next step to be solved to facilitate the adoption of fog technologies.

The optimization of resource management in fog computing architectures has two main issues: service placement and data allocation. Our proposal is addressed to optimize service placement. Previous fog service management algorithms have explored a wide range of optimization techniques, such as heuristics, greedy algorithms, linear programming, or genetic algorithms, among others. These service managers have defined several aspects of the fog resources, such as placement, scheduling, allocation, provisioning, or mapping for services, resources, clients, tasks, virtual machines, or even fog colonies.  These solutions have been defined for environments such as industrial IoT, smart cities, eHealth, or mobile users.

Table~\ref{survey1} presents a brief survey of optimization proposals for the FSPP. This problem deals with the idea of placing cloud services closer to the clients, by using the computational resources of networking and edge components. FSPP differs from computational offloading problem~\cite{Kumar2013,7879258} in that the latter migrates the services of the client's devices to the cloud or to the edge. Offloaded services are particular of each user, in contrast to the FSPP where services are shared between clients. Another difference is that offloading optimization does not deal with service scalability, as the FSPP does. We do not include researches in this latter field due to the differences with FSPP. Data allocation is also a current research topic in fog resource management, but the nature of data and services is different enough to be necessary to propose particular solutions for their management. Service offloading and data allocation are out of the scope of FSPP, so they are not included in this brief survey.

The characteristics of the related work have been summarized in Table~\ref{survey1}, by indicating the optimization purpose (column \textit{Objective functions}), the elements that the optimization algorithm manages to improve the objective functions (column \textit{Decision variables}), and the optimization algorithm (column \textit{Algorithm}).

\begin{table}[t!]
	\centering
	\caption{Summary of Fog Service Placement Problem studies.}
\label{survey1}
\scriptsize
			\begin{tabular}{p{0.03\columnwidth}p{0.3\columnwidth}p{0.35\columnwidth}p{0.2\columnwidth}}		
\toprule	
	\textbf{Ref.} 
	& \textbf{Objective functions}  & \textbf{Decision variables}  
	&  \textbf{Algorithm}

	\\
	
	\midrule

	\cite{7867735} &  Response time, QoS & Service placement &  GA \\

	\cite{Skarlat2017} &  Resource waste, Execution times & Fog colony service placement & GA, First fit \\

	\cite{7110527} &  Cost, Latency and Migration & Service placement and Load dispatching &  Greedy, ILP, GA \\

	\cite{ARKIAN2017152}  &Cost  & Client association, Resource provisioning, Task distribution, VM placement&  Mixed ILP \\
	
	\cite{7359164} &  Cost & Base station association, Task distribution, VM placement &  ILP \\
	
	\cite{Velasquez2017} &  Network latency, Service migrations & Service placement &  ILP \\
	
	\cite{HUANG201447}&  Communication power consumption & Service merging and placement  & ILP, Maximum Weighted Independent Set \\
	
	\cite{7511465} &  Service delay & Service placement &  ILP \\
	
	\cite{8014364} &  Deadline violations, Cost, Response time& Service placement &  ILP \\
	
	\cite{7422054}&  Task completion time & Task scheduling, Task image placement, Workload balancing &  Mixed ILP \\
	
	\cite{7676307} &  Power consumption & Service placement &  ILP \\

	\cite{7935527} & Response time, Cost & Resource allocation and scheduling  &  Petri Nets\\
	
	\cite{URGAONKAR2015205} &  Queue length, cost & Service migration & Decoupled Markov Decision Process\\
	
	\cite{8014366} &  Resource consumption, QoS & Service placement &  MonteCarlo \\

	\cite{8588297} & Availability, QoS & Service placement & Complex networks \\
	
	\cite{COLISTRA201498} &  Resource usage& Resource allocation &  Consensus \\
	
	\cite{7249199} &  Cost & Look-ahead service placement  & Shortest Path \\
	
	\cite{Urgaonkar:2015:DSM:2822545.2822799} &  Cost & Look-ahead service placement  & Markov Decision Process \\

	\cite{7035717} & Cost  & Task placement &  Binary prog., Greedy\\ 
	
		\cite{SPE:SPE2509}& Energy, Network usage, Latency & Service placement &  First Fit\\
	
	\cite{7987464}  &  Network usage, Power consumption, Latency & Service placement &  Own algorithm\\
	
	\cite{7847322} &  Load balancing & Service placement &  Own algorithm \\
	
	\cite{7912261} &  Network usage and delay & Service placement &  Own algorithm \\
	
	\cite{7389120} &  Executed tasks & Resource provisioning& Own algorithm\\
	
	\cite{7248934} &  Power consumption, Response time & Workload placement &  Own algorithm\\
	
	\cite{Saurez2016}  & Latency & Service placement and migration &  Own algorithm\\

	\cite{7917628}& Network latency & Task assignment & Own algorithm\\
	
	\cite{Mahmud2018} & 	Service latency, QoS &  Service placement & Own algorithm \\

	\\
	
	[This work] &  Resource usage, network latency, service spread & Service placement &  WSGA, NSGA-II, MOEA/D \\
	
\bottomrule	
\end{tabular}
\end{table}

Since our main contribution is the study of the applicability of three genetic algorithms to the FSPP, we group previous research works by the optimization algorithm they implemented. Additionally, we focus our attention in research where evolutionary or genetic approaches were considered. The other researches are only listed in the related work. 

Probably, Integer Linear Programming (ILP) is the most common solution in the field. Up to 9 papers used ILP to optimize cost~\cite{7110527,ARKIAN2017152,7359164,8014364}, latency or execution times~\cite{7110527,Velasquez2017,7511465,8014364,7422054} , migrations~\cite{7110527,Velasquez2017}, QoS~\cite{8014364} or power consumption~\cite{HUANG201447,7676307}. 

Other types of algorithms are less representative and a smaller number of studies have investigated their implementation in the field of fog computing. These are for example greedy algorithms~\cite{7110527,7035717,SPE:SPE2509}, Markov Decision Process~\cite{Urgaonkar:2015:DSM:2822545.2822799,URGAONKAR2015205}, Petri Nets~\cite{7935527}, Monte Carlo~\cite{8014366}, complex networks~\cite{8588297}, consensus~\cite{COLISTRA201498}, or shortest path~\cite{7249199}.

Moreover, several authors have explored new algorithms that were not based on standardized ones. Some of those algorithms are based on:  a sequentially assignment of the highest demanding application modules to the nodes with biggest capacities~\cite{7987464}; the use of a service placement with likely performance bounds and a linear graph~\cite{7847322}; on the mobility of the users, the fog proximity and the elasticity of the cloud~\cite{7912261}; the decomposition into subproblems~\cite{7389120,7248934}; a mobility-based algorithm~\cite{Saurez2016}; a future load estimator~\cite{7917628}; a constrained-based forwarding algorithm~\cite{Mahmud2018}.

To the best of our knowledge, there are only three previous studies that considered genetic algorithms for the optimization of the FSPP. Firstly, Wen et al.~\cite{7867735} presented a preliminary proposal based on a parallel GA to reduce response time and increase the QoS. They compared a parallel GA and a serial one, and they only tested a weighted sum optimization. They did not provide details of the implementation and the decision variables. Yang et al.~\cite{7110527} provided a deeper analysis of the use of a GA and they compared it with a greedy heuristic and an ILP-based solution. But they only incorporated one optimization objective in the study. Finally, Sarklat et al.~\cite{Skarlat2017} proposed a single objective GA by transforming the two considered objectives with a weighted sum function. They organized the fog devices into colonies with a coordinator device and the optimization algorithm decides whether the services are placed inside the colonies or they are propagated to neighboring colonies.

In summary, previous approaches are mainly addressed to optimize a single objective and, in the case of the ones based on genetic algorithms, they are implemented with simple approaches. Additionally, none of them addressed the scalability of the services. To the best of our knowledge, our approach is the first one which addresses a multi-objective optimization of the fog service placement problem, considering a scalable service replication level, and using pure multi-objective genetic algorithms, such as NSGA-II or MOEA/D.

\section{Problem statement}
\label{problemstatement}

Fog computing is an architecture that distributes the computing and storage functions of traditional cloud-based applications to devices closer to the users along a cloud-to-thing continuum. These devices with computational and storage capabilities, commonly called fog devices, are distributed across the layers of the network topology~\cite{openfog}. Data and service management policies are needed to decide when and where to place the services and the data. The problem of the selection of the fog devices where the services are instantiated is usually named as FSPP. In the remaining of this section, we formally define our domain for the FSPP.

We focus our work in the study of the algorithms that decide the placement of the services in the fog devices. Certainly, the solutions to the FSPP are not adaptive and fog domains are environments with changing conditions. But these challenges have been already addressed in some previous works that have proposed frameworks for the integration of resource management policies in real fog environments~\cite{donassolo:hal-01859695,8603163,Velasquez2017}. For example, any of our three proposed algorithms could be integrated in the framework in Velasquez et al.~\cite{Velasquez2017}. In this framework, the changing conditions in the architecture are gathered by the \textit{Information Collection} module. The \textit{Service Orchestrator} module uses these data of the state of the system to take decisions about the service placements. Our algorithms would be integrated in this \textit{Service Orchestrator} module.
 
\subsection{System model}
\label{systemmodel}

\begin{figure}[t]
	\includegraphics[width=0.9\linewidth]{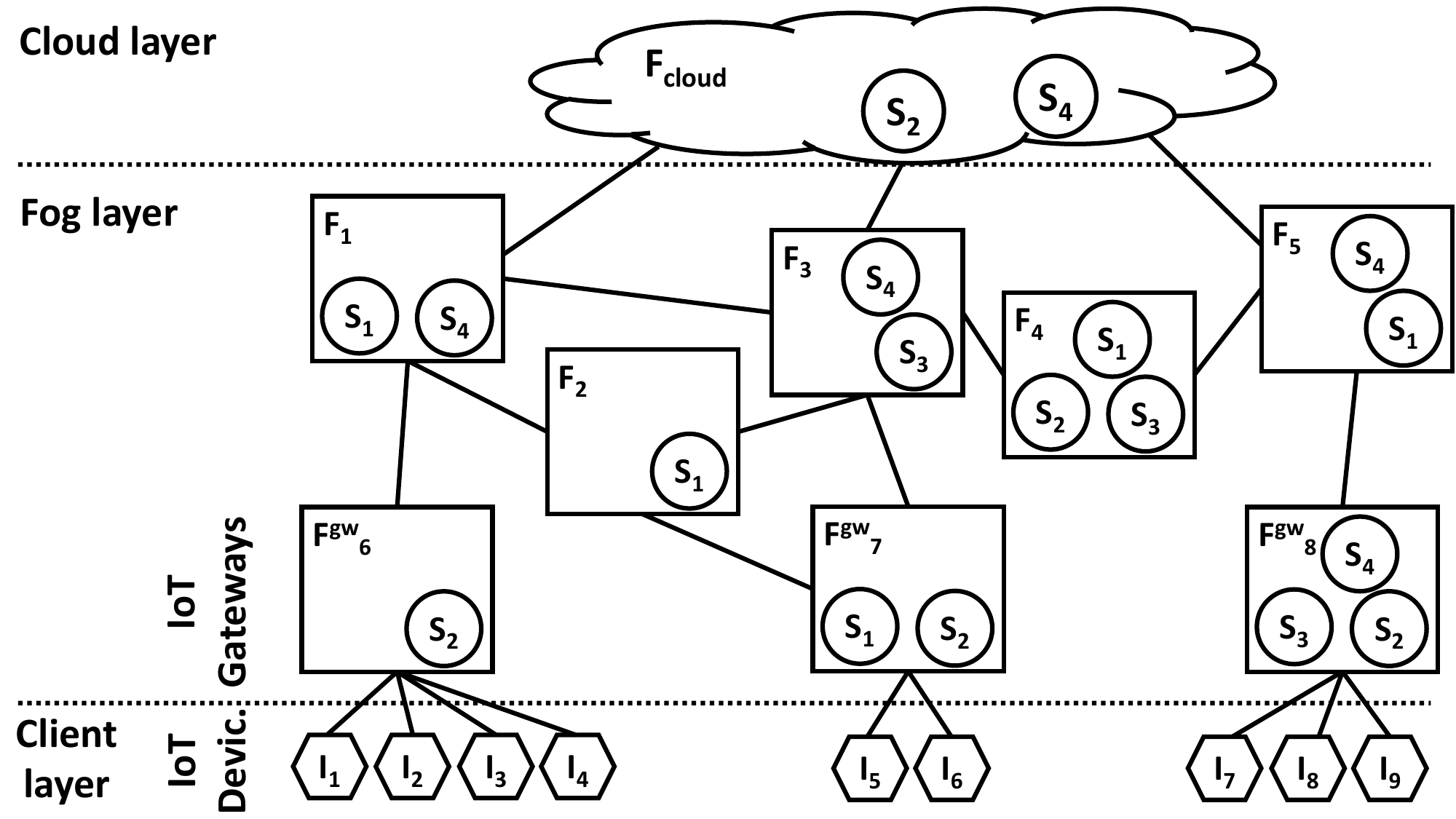}
	\caption{An example of a fog computing architecture.}
	\label{fogarchitecture}
\end{figure}

An example of a fog computing architecture is represented in Figure~\ref{fogarchitecture} where three layers can be identified: cloud layer, fog layer, and client or device layer. The cloud layer corresponds to the cloud provider, where services can (or not) be stored and executed, as in any other device in the system. The client layer is typically formed by the IoT devices (sensor and actuators) that generate data to be stored, request services, or consume data. They request the services and receive the responses. The fog layer includes the in-network intermediate devices that are able to execute instances of the services. We assume that the connection between the fog devices is a network with a graph structure. We also assume that the IoT devices can be only connected to special devices that have both roles, fog devices, and IoT gateways.

Several models for fog applications have been defined. One of the most popular developing patterns for IoT applications in the fog is the distributed dataflow~\cite{7356560}. But the use of microservice-based applications has been also proposed for the deployment of fog applications~\cite{7436659,Vogler:2016:SFP:2909066.2850416,Saurez2016,7522321} making easier the scale up and down of service instances when stateless and decoupling are guaranteed~\cite{7796008}. They both are based on the definition of the applications as a set of services that interoperate between them by sending messages. Both models separate the management of the data (storage) and the computation (services), making even easier the management and scaling of the applications~\cite{7945515,6676764,7868163,7300793}.

We model the architecture as a graph structure where the nodes $F=\{f_i\}$ are the fog devices and the edges $C=\{c_{i,i'}\}$ the direct network links between the fog devices. There are two special types of devices, $f^{cloud}$ that represents the cloud provider, and $GW=\{f_i^{gw}\}$ the subset of fog devices that also act as IoT gateways. The devices are characterized by the parameter $R_{f_i}^{cap}$ that is the total resource capacity of the device. We assume that the resources in the cloud provider are infinite, $R_{f^{cloud}}^{cap}=\infty$. In order to simplify the model and the notation, we assume a scalar value for the computational resource capacities. It can easily be extended using, for example, a tuple for each resource element (e.g., CPU capacity, main memory size, storage size, or input/output bandwidth). The network connection links are characterized by the communication latency, $\mathit{L}_{c_{i,i'}}$.

The applications are modeled as a directed graph where the nodes $S=\{s_x\}$ are the services that are related through many-to-many consumption relationships. This relationship is represented with a matrix $I$ of size $|S| \times |S|$, where $|S|$ is the number of services, and its element $\iota_{x,x'}$ are equal 1 if $s_x$ consumes $s_{x'}$, and 0 otherwise.  The services can be scaled up or down and, consequently, several instances of the services can be deployed in different devices. These instances are identified with $s_x^y$. The services are characterized by their resource consumption, $R^{con}_{s_x}$.

The allocation of the service instances in the fog devices is represented with a matrix $A$ of size $|S| \times |F|$ where $|S|$ is the number of services and $|F|$ the number of devices. The element $\alpha_{x,i}$ is equal 1 if the fog device $f_i$ hosts an instance of the service $s_x$, and 0 otherwise.

The IoT devices (\textit{things}), typically sensors and actuators, are represented as $T=\{t_n\}$, and they are characterized by the services they request and the IoT gateway they are connected to. Consequently, a matrix $R$ of size $|GW| \times |T|$ is defined to represent the origin of the application request, where $|GW|$ is the number of gateways, $|T|$ the number of IoT devices, and $\rho_{i,n}$ is equal 1 if IoT gateway $f_i^{gw}$ has at least one IoT device $t_n$ that requests the service $s_x$ associated with the IoT device, and 0 otherwise. To simplify the notation, we define $\rho_{i,x}=1$ to represent that the gateway $f_i^{gw}$ requests the service $s_x$.

We define the symmetric matrix of shortest latency distances between nodes, $ D = |F| \times |F|$, calculated as the summation of the link latencies within the shortest path between a pair of nodes:

\begin{equation}
d_{i,i'} = d_{i',i} = \sum^{\forall\ c_{q,r} \in shortestPath(f_i,{f_{i'}})} \mathit{L}_{c_{q,r}} 
\end{equation}

\begin{table}[t]
	\centering
		\caption{Summary of the variables of the system model.}%
\label{systemmodel}

		\begin{tabular}{lp{0.8\columnwidth}}
\toprule
\textbf{Variable} & \textbf{Description}  \\
\midrule
$S$ & Set of the services in the system\\
$s_x$ &  A service in the system\\
$s^y_x$ &  An instance of a service $s_x$ \\
$\iota_{x,x'}$ &  Element of matrix $I$ that indicates if $s_x$ requests $s_{x'}$\\
$F$ & Set of the fog devices in the system\\
$f_i$ &  A fog device in the system\\
$f^{cloud}$ & Identification of the cloud provider\\
$f_i^{gw}$ & Identification of an IoT gateway\\
$R^{cap}_{f_i}$ & Resource capacities of a device $f_i$\\
$R^{con}_{s_x}$ & Resource consumption required by a service $s_x$\\
$\alpha_{x,i}$ & Element of the matrix $A$ that indicates if $s_x$ is allocated in $f_i$\\
$T$ & Set of the IoT devices in the system\\
$t_n$ &  An IoT device in the system\\
$\rho_{i,n}$ / $\rho_{i,x}$ & Element of matrix $R$ that indicates if the IoT gateway $f_i^{gw}$ has an IoT device $t_n$ that requests service $s_x$.\\
$c_{i,i'}$ & Direct network connection link between devices $f_i$ and $f_{i'}$\\
$\mathit{L}_{c_{i,i'}}$ & Communication latency of a link $c_{i,i'}$,  i.e., the communication latency between two directly connected devices.\\
$d_{i,i'} $ & Communication latency between any two devices $f_i$ and $f_{i'}$\\

\bottomrule
\end{tabular}

\end{table}

\subsection{Optimization model}

Our main concerns about the system are three: (i) to maximize the use of the fog devices to take as much advantage as possible of placing the services closer to the clients; (ii) to evenly distribute the service instance replicas across the fog devices; and (iii) to reduce the network latency due to network communications. The following paragraphs explain three metrics that we selected as indicators of these there optimization objectives.

\subsubsection{Resource usage}

We include this objective to maximize the use of the resources in the fog devices. The maximization of the resource usage is based on the idea that it is better to deploy as many services as possible in the fog devices. When there are many services deployed in the fog devices, the probability of a user to request a service that is allocated closer than the cloud provider is increased. Consequently, in the cases that the fog devices still have free resources, the placement could probably be improved by placing more services in these free resources. Under these conditions, the more the usage of the fog devices and their resources is, the better the solutions are. We assume that the optimal service placement would be to place instances of each service in all the fog devices. But this is impossible due to the constraint of unlimited resources in the devices. But we desire that, at least, all the services are used up to the 100\%. This does not damage the flexibility of the system, because we assume that the removal of services is as simple as stopping the service and deleting it since we consider stateless microservice-based applications~\cite{7945515,6676764}.

Therefore, our first optimization objective is to minimize the free resources in the devices. We calculate the resource usage as the ratio between the resource consumed by the instances of the services and the total available resources:
\begin{equation}
Resource\ Usage = \frac{\sum_{f_i}^{F} \sum_{s_x}^{S} \alpha_{x,i} \times R_{s_x}^{con}}{\sum_{f_i}^{F} R_{f_i}^{cap}} 
\end{equation}
where the numerator is an iteration through the elements in the matrix, summing the resource consumption for the cases that the instances are allocated ($\alpha_{x,i}=1$). The denominator is the summation of the capacities of all the devices.

Since we focused the optimization in a minimization process, we used the metric that represents the free resources instead of the resource usage, defined as:

\begin{equation}
Free\ Resources = 1.0 - Resource\ Usage
\end{equation}

\subsubsection{Service spread}

Our second objective is to evenly distribute the service replicas across the fog domain to increase the coverage for all the users. Note that having the replicas of services concentrated in closer devices affects: (a) The latencies of the furthest users, which could be reduced by moving some instances away; (b) The service availability because the concentrated services could get isolated in a region of the network if some links fail. If the services are spread, this isolation is more unlikely. 

We used the coefficient of variation (CV), or  relative standard deviation, of the network latencies between each pair of replicas to measure the distribution of the service replicas. CV is calculated as the ratio between the standard deviation and the mean of a data series. The value of this measure is nearing to zero as the elements get more dispersed. We defined the service spread as the average of the single CV of each service:
\begin{equation}
Service\ Spread=\frac{\sum_{s_x}^{S}\frac{\ \sigma^{d_{s_x}^{rep}}\ }{\ \overline{d_{s_x}^{rep}}\ }}{|S|}
\end{equation}
where $\overline{d_{s_x}^{rep}}$ and $\sigma^{d_{s_x}^{rep}}$ are respectively the average and the standard deviation of the network distances between each pair of replicas $s_x^y$ of a service $s_x$. 

The average, $\overline{d_{s_x}^{rep}}$, is calculated as:
\begin{equation}
\overline{d_{s_x}^{rep}} = \frac{\sum_{i=0}^{F} \sum_{i'=i+1}^{F} \alpha_{x,i} \times \alpha_{x,i'} \times d_{i,i'}}{\sum_{i=0}^{F} \sum_{i'=i+1}^{F} \alpha_{x,i} \times \alpha_{x,i'}}
\end{equation}
where the numerator goes through the devices that allocates one service, and adds the distances of each pair of instances of that service. The denominator calculates the number of pairs of instances of that service.

The standard deviation, $\sigma^{d_{s_x}^{rep}}$, is calculated as:
\begin{equation}
\sigma^{d_{s_x}^{rep}} = \sqrt{\frac{\sum_{i=0}^{F} \sum_{i'=i+1}^{F} \alpha_{x,i} \times \alpha_{x,i'} \times (d_{i,i'}-\overline{d_{s_x}^{rep}})^2}{\sum_{i=0}^{F} \sum_{i'=i+1}^{F} \alpha_{x,i} \times \alpha_{x,i'}}}
\end{equation}
where, as in the case of the average, the numerator goes through each pair of instances of a service, and the denominator is the number of pairs of instances.

\subsubsection{Network latency}

One of the main reasons to use fog computing is to reduce the network latency between the users and the services. We have also included that in our optimization objectives, but considering the network distances between all the services in an application, and not only the distance between the users (IoT devices) and the first service.

We defined a metric based on the distances between a service instance and the closest instance of each of their consumed services. We also considered the specific case of the distance between the IoT devices (connected to the IoT gateways) and their requested services. We defined an indicator of the network latency as the average value of the distances between interoperated services ($d_{s_x}^{cons}$) and between IoT devices and requested services ($d_{t_n}^{req}$):
\begin{equation}
Network\ Latency = \frac{\sum_{s_x}^{S} d_{s_x}^{cons} + \sum_{f_i}^{GW} d_{f_i^{gw}}^{req}}{|S| + |GW|}
\end{equation}
where $|GW|$ is the number of IoT gateways, and $|S|$ the number of services as it has been previously explained.

The distance of interoperated services $d_{s_x}^{cons}$ for a service $s_x$ is calculated as the average value of the minimum distance between each of its instances and the closest instance of all its consumed services:
\begin{equation}
d_{s_x}^{cons} =  \frac{\sum_{f_i}^{F}\sum_{s_{x'}}^{S}\alpha_{x,i} \times \iota_{x,x'} \times \min_{i'=0}^{F}(d_{i,i'}\times \alpha_{x',i'})}{\sum_{f_i}^{F}\sum_{s_{x'}}^{S}\alpha_{x,i} \times \iota_{x,x'}}
\end{equation}
where $x$ is the index of the service whose distance is calculated for, $i$ the indices of the devices which allocate its instances, $x'$ are the indices of their consumed services, and $i'$ the indices of the devices that allocate the instances of the consumed services. The numerator goes through the allocation of each instance of a service (first summation), through the services consumed by that services (second summation), and takes the minimum distance between the device of the the current instance and all the devices allocating the consumed service (calculated with $min_{i'=0}^{F}$). 

The distance between IoT devices and requested services is defined similarly to the previous one but the origin is an IoT gateway, instead of a service, and the targets are the services requested from the clients connected to the gateway. The formula is:
\begin{equation}
d_{f_i^{gw}}^{req} =  \frac{\sum_{s_x}^{S} \rho_{i,x} \times \min_{i'=0}^{F}(d_{i,i'}\times \alpha_{x,i'})}{\sum_{s_x}^{S}\rho_{i,x}}, i \neq i'
\end{equation}
where $i$ is the index of the IoT gateway, $x$ the indices of the requested services by the clients connected to the IoT gateway, and $i'$ are the indices of the devices that allocate the instances of the requested services.

\subsubsection{Model constraints}

Our model only has one constraint. The resources consumed by the services in the fog devices have to be lower than the resource capacity of these devices. Constraints in the resource usages of the network links are not considered, but they could be included in future works. Consequently, the following constraint needs to be accomplished:

\begin{equation}
\label{usageconstraint}
\sum_{s_x}^{S}\alpha_{x,i} \times R^{con}_{s_x} \leq R^{cap}_{f_i}, \forall\ f_i\ \in\ F
\end{equation}

To sum up, our optimization is addressed to minimize $Free\ Resources $ $\land$ $Service\ Spread$ $\land$ $Network\ Latency$, by determining the values $a_{x,i}$ of the allocation matrix $A$ subject to constraint in Eq.~\ref{usageconstraint}.


\section{Evolutionary optimization}
\label{evolutionary}

Computational Intelligence (CI) is a common solution for resource management problems in the field of cloud resource management~\cite{guzek2015survey}. Evolutionary algorithms (EA), along with particle swarm, neural networks or fuzzy systems, are CI techniques commonly used~\cite{zhan2015cloud}. The adaptation of those techniques to each particular problem makes necessary to perform a specific study for each new particular scenario~\cite{wolpert97nofree}.

We have studied the suitability of three evolutionary algorithms to solve the FSPP: the weighted sum genetic algorithm (WSGA), the non-dominated sorting genetic algorithm-II (NSGA-II), and the multiobjective evolutionary algorithm based on decomposition (MOEA/D). Studies of the general performance of those three algorithms have been previously presented~\cite{4358754} but, to the best of our knowledge, they have not been studied for the domain of fog architectures. Our contribution is to study the suitability of these three algorithms in fog domains.

Additional to the general guidelines of each of the three algorithms, explained in the next section, our approach requires of a common definition of some elements and operators, such as the solution representation, or the crossover and mutation operators, explained in Section~\ref{geneticoperators}.

\subsection{Evolutionary algorithms}

The three algorithms of our study are inspired by the biological evolution. A population of solutions is evolved along generations by combining and changing them with the use of crossover and mutation operators. A fitness function, based on the objectives to be optimized, defines the quality of the solutions, which also determines the probability of being mated.

\subsubsection{Weighted sum genetic algorithm}

GAs with single objective optimization are one of the preliminary solutions for EA. These algorithms use the scalar value of the objective function as the fitness value for each solution. In the case of multi-objective optimization, it is necessary to establish a  linear transformation of the multi-objective into a scalar value.

The first algorithm of our approach is a single objective GA based on the use of a weighted sum transformation (WSGA). This transformation consists on normalizing the values of the objectives to the unit interval, applying a weight, and summing them:
\begin{equation}
\label{eqgenweighted}
\sum_{i}^{num.\ objectives}\omega_{i} \times \theta_{i} \times x_i 
\end{equation}
where $\omega_{i}$ is the scaling factor, $\theta_{i}$ the weight, and $x$ the value of the objective function.

Algorithm~\ref{weightedsum} shows the general structure of the WSGA. The algorithm starts generating $populationSize$ random solutions, whose objective values are calculated and weighted sum transformed. Along $generationNumber$ iterations of the algorithm, $populationSize$ children are generated in each iteration.

The children solutions are created by applying a crossover operator over two father solutions of the previous population. The outputs of a crossover operation are two children. These children are mutated with a probability of $mutationProb$. The details of both operators are in Section~\ref{geneticoperators}.

The fathers of a crossover operation are chosen using a deterministic binary tournament selection operator~\cite{GOLDBERG199169}. First, two solutions are chosen from the population, and the one with the best fitness is selected as the first father. This is repeated again for the second father.
 
Children only replace former solutions if they have better fitness values. This is implemented by creating a population with the previous population and the offspring, ordering them by their fitness values, and creating the new population with the first $populationSize$ solutions with the highest fitness. After the $generationNumber$ iteration, the solution with the best fitness value is selected as the output of the algorithm.

\begin{algorithm}[t]
	\caption{Single-objective traditionally genetic optimization algorithm}
	\label{weightedsum}
	\begin{algorithmic}[1]
		\Procedure{WSGA}{}
		\State $P_t \gets generateRandomPopulation(populationSize)$ \label{alg_rand_gen}
		\State $objectiveValues \gets evaluateObjectiveFunctions(P_t)$
		\State $fitness \gets weightedsum(objectiveValues,\omega, \theta)$
		\For{$i\ $in$\ 1..generationNumber$}
		\State $P_{off} = \emptyset$ 
		\For{$j\ $in$\ 1..populationSize$} \label{alg_beg_for}
		\label{alg_sel1}
		\State $father1 \gets binaryTournament(P_t,fitness)$
		\State $father2 \gets binaryTournament(P_t,fitness)$ 
		\label{alg_sel2}
		\State $child1,child2 \gets crossover(father1,father2)$
		\If {$random() < mutationProb$} 
		\State $mutate(child1)$,$mutate(child2)$
		\EndIf
		\State $P_{off} = P_{off} \cup \{child1, child2\} $ 
		\EndFor \label{alg_end_for}
		\State $objectiveValues \gets evaluateObjectiveFunctions(P_{off})$
		\State $fitnessOff \gets weightedsum(objectiveValues,\omega, \theta)$
		\State $fitness = fitness \cup fitnessOff$ 
		\State $P_{off} = P_{off} \cup P_{t}$  \label{alg_join}
		\State $P_{off} = orderElements(P_{off},fitness)$ \label{alg_order}
		\State $P_{t} = P_{off}[1..populationSize]$  \label{alg_half}
		
		\EndFor
		\State $Solution = min(P_{t},fitness)$
		
		\EndProcedure
	\end{algorithmic}
\end{algorithm}

\subsubsection{Non-dominated sorting genetic algorithm-II }

The second algorithm of our study, the non-dominated sorting genetic algorithm-II (NSGA-II)~\cite{deb2002fast} orders the solutions using the dominance concept. A transcription of the original algorithm is shown in Algorithm~\ref{nsga2}.

Multi-objective optimization algorithms introduce the concept of dominance to order the solutions, instead of using a scalar value. A solution $s_1$ dominates another solution $s_2$ if all the objectives values of $s_1$ are better than the values of $s_2$. In the same manner, a solution $s'_1$ non-dominates another solution $s'_2$, if $s'_2$ has al least one objective with a better value. Consequently, the solutions in a population can be classified in dominating and dominated solutions. The Pareto optimal front is defined as the set of solutions that are non-dominated by any other solution. Therefore, each solution in the Pareto front optimizes one or more objectives, but not all of them, in regard to the other solutions in the front. The solution of a multi-objective optimization is a set of solutions, the Pareto set, instead of a single solution.

The NSGA-II mainly differs from the WSGA in how the fitness is represented and how the solutions are ordered. The fitness is a vector with one element for the value of the objective functions. The solutions are ordered in successive fronts. The Pareto optimal front is the first one. Once that the Pareto optimal front is calculated, the remaining solutions are again processed to calculate a new front without considering the solutions already placed in the previous one. This is repeated until all the solutions are included in one front. The solutions inside a front are ordered with the crowding distance, that is calculated as the minimum Euclidean distance from one solution to the others. The NSGA-II considers that dispersed solutions are more significant than the solutions that are concentrated together. The details of the algorithms are deeply explained in the study of its first proposal~\cite{deb2002fast}. The fathers are chosen with a binary tournament selector, like in the WSGA, but instead of comparing the scalar fitness value, the front and the crowding distance are considered. 

The output of the algorithm is the Pareto optimal front of the last iteration of the algorithm.

\begin{algorithm}[t]
	\caption{Multi-objective genetic optimization algorithm~\cite{deb2002fast}}
	\label{nsga2}
	\begin{algorithmic}[1]
		\Procedure{NSGA-II}{}
		\State $P_t \gets generateRandomPopulation(populationSize)$ \label{alg_rand_gen}
		\State $fitness \gets calculateFitness(P_t)$
		\State $fronts \gets calculateFronts(P_t,fitness)$
		\State $distances \gets calculateCrowding(P_t,fronts,fitness)$
		\For{$i\ $in$\ 1..generationNumber$}
		\State $P_{off} = \emptyset$ 
		\For{$j\ $in$\ 1..populationSize$}
		\label{alg_sel1}
		\State $father1 \gets binaryTournament(P_t,fronts,distances)$
		\State $father2 \gets binaryTournament(P_t,fronts,distances)$ 
		\label{alg_sel2}
		\State $child1,child2 \gets crossover(father1,father2)$
		\If {$random() < mutationProb$} 
		\State $mutate(child1)$,$mutate(child2)$
		\EndIf
		\State $P_{off} = P_{off} \cup \{child1, child2\} $ 
		\EndFor \label{alg_end_for}
		\State $fitness \gets calculateFitness(P_{off})$ \label{alg_front}
		\State $P_{off} = P_{off} \cup P_{t}$  \label{alg_join}
		\State $fronts \gets calculateFronts(P_{off},fitness)$ 
		\State $distances \gets calculateCrowding(P_{off},fronts,fitness)$
		\State $P_{off} = orderElements(P_{off},fronts,distances)$ \label{alg_order}
		\State $P_{t} = P_{off}[1..populationSize]$  \label{alg_half}
		
		\EndFor
		\State $Solution = fronts[1]$ \#the Pareto front
		
		\EndProcedure
	\end{algorithmic}
\end{algorithm}

\subsubsection{Multi-objective evolutionary algorithm based on decomposition}

Finally, the third algorithm of our study is the multi-objective evolutionary algorithm based on decomposition (MOEA/D)~\cite{4358754}. This algorithm is based on decomposing the problem into $N$ scalar fitness optimizations that are simultaneously optimized along the iterations. A transcription of the original algorithm is shown in Algorithm~\ref{moead}.

At the beginning of the algorithm, $N$ vectors of weights are generated evenly distributed, along with a random solution for each of these weight transformations. In each iteration, a solution for each $N$ weight vector is attempted to be optimized. The algorithm is based on the idea that a solution has the same quality for all the neighboring weight vectors. The neighboring vectors are measured in terms of the euclidean distance of the weights. By this, the fathers of a new solution for a given vector are randomly chosen from the solutions of the $T$ closest weight vectors. The two children of the crossover are compared and the dominating one is selected. The current solution for the weight vector is only replaced by the new child if the latter dominates the former. Additionally, an external population ($EP$) is considered, where the child is included when it is not dominated by any other solution currently in the EP. If the child is included, all the solutions that it dominates are removed from the EP.

\begin{algorithm}[!t]
	\caption{Multi-objective evolutionary algorithm~\cite{4358754}}
	\label{moead}
	\begin{algorithmic}[1]
		\Procedure{MOEA/D}{}
		\State $P_{EP} \gets \{\}$
		\State $W_N \gets generateEvenlyWeights(N)$
		\State $P_N \gets generateRandomPopulation(N)$
		\For{$j\ $in$\ 1..N$}
		\State $B[j] \gets getClosestWeights(T)$
		\EndFor
		\For{$i\ $in$\ 1..generationNumber$}
		\State $P_{off} = \emptyset$ 
		\For{$j\ $in$\ 1..N$}
		\State $father1 \gets B[j][rand(1,T)]$
		\State $father2 \gets B[j][rand(1,T)]$
		\State $child1,child2 \gets crossover(father1,father2)$
		\If {$rand() < mutateProb$} 
		\State $mutate(child1)$,$mutate(child2)$
				\EndIf
		\State $child \gets dominant(child1,child2)$
		\For{$k\ $in$\ 1..T$}
		\If {$child\ dominates\ P_N[B[j][k]]$} 
		\State $P_N[B[j][k]] \gets child$
		\EndIf
		\EndFor
		\State $P_{off} = P_{off} \cup \{child\} $ 
		\EndFor \label{alg_end_for}
		\State $fitness \gets calculateFitness(P_{off})$ \label{alg_front}
		\State $P_{off} = P_{off} \cup P_{EP}$  \label{alg_join}
		\State $fronts \gets calculateFronts(P_{off},fitness)$ 
		\State $P_{EP} \gets fronts[1]$
		
		\EndFor
		\State $Solution = P_{EP}$ \#the Pareto front
		
		\EndProcedure
	\end{algorithmic}
\end{algorithm}

\subsection{Genetic operators and structures}
\label{geneticoperators}

The three algorithms of the previous section use the same solution representation, and genetic operators (crossover and mutation). The following paragraphs explain the details.

The solutions of our optimization, also known as individual or chromosome in the field of evolutionary algorithms, are the allocations of the fog services in the fog devices. If we consider the model of Section~\ref{systemmodel}, a solution is represented by the matrix $A$.

In evolutionary algorithms, solutions are usually represented with n-dimensional array structures which represent the chromosome solution and each element of the array is usually known as gene~\cite{gen2000genetic}. Consequently, our solutions are two-dimensional arrays that directly represent the matrix $A$. The rows are the fog services and the columns the fog devices. The value of an array element is 1 when the service is allocated in the device and 0 otherwise.

Evolutionary algorithms generate new solutions by the combination of the current best solutions based on the biological concept of evolution~\cite{srinivas1994adaptive}. This combination is performed with the crossover operator, obtaining two new solutions that alternative take array elements (genes) from both fathers. We selected a single-point crossover operator~\cite{mitchell1998introduction}. A different random number $r$, between 1 and the number of devices, is generated for each service allocation (for each row). This random number splits the solution row into two pieces in both fathers. The opposite pieces from each father are combined, the first child is obtained by concatenating the $[1,r]$ elements of the first father and the $[r+1, \#devices]$ of the second one. The second child is obtained from the other two opposite pieces.

The population evolution based only on crossover operations has the risk of falling into local optimizations. The mutation operator is included in evolutionary algorithms to expand the solution search space and to avoid local minimums~\cite{mitchell1998introduction}. We defined three mutation operators: \textit{replica growth}, that randomly increases the number of instances of each service; \textit{service shuffle}, that interchanges the allocation plan of all the services in the system; \textit{spread to fog}, that randomly selects a subset of services and they are instantiated and allocated in all the fog devices. 

It is very likely to obtain an allocation plan that does not satisfy the constraints of our model, mainly with very disruptive mutations such as the last one. If a solution does not satisfy the resource usage constraint (Eq.~\ref{usageconstraint}) after a crossover or mutation, it is modified with a mend operator. This operator is an iterative process of removing random instances of services from the fog devices that do not satisfy the constraint, until the resources consumed by the allocated services are smaller or equal to the device capacity.

\section{Experimental validation}
\label{sect:expvalidation}

For the study and the validation of the three evolutionary algorithms, we first defined an experiment with a common infrastructure, a set of applications, and the user characteristics (Section~\ref{sect:expdefinition}). Additionally, the parametrization of the evolutionary algorithm was fixed in a preliminary and exploratory phase where several alternatives were studied and the most suitable values were selected (Section~\ref{sectevolalgparameters}).

\subsection{Experiment definition}
\label{sect:expdefinition}

The experiments were defined in terms of the fog device features, network topology, application characteristics and clients' distribution. The experiments were defined having in mind a potential model of a region of a bigger fog computing architecture. In any case, the sizes of our experiments are bigger than most of the experiments performed in the related bibliography (Section~\ref{relatedsect}). We considered 100 fog devices, 100 and 200 services, and 8 users/IoT devices per gateway (resulting in a total number of 160 users).

We random generated the topology of the network as an Albert-Barabási topology with 100 devices. This is a common model for autonomous network topology that has been also used in other fog resource management studies~\cite{8368525}. The cloud provider was placed in the node with the highest betweenness centrality.  The nodes with the smallest centrality were designated as IoT gateways. Betweenness centrality is a graph metric that quantifies the number of times a node acts as a bridge along the shortest path between two other nodes~\cite{Koschuzki2005}. It is usually considered as an indicator of the control over the communications between any two nodes.

\begin{figure}[t]
	\center
	\includegraphics[width=0.9\linewidth]{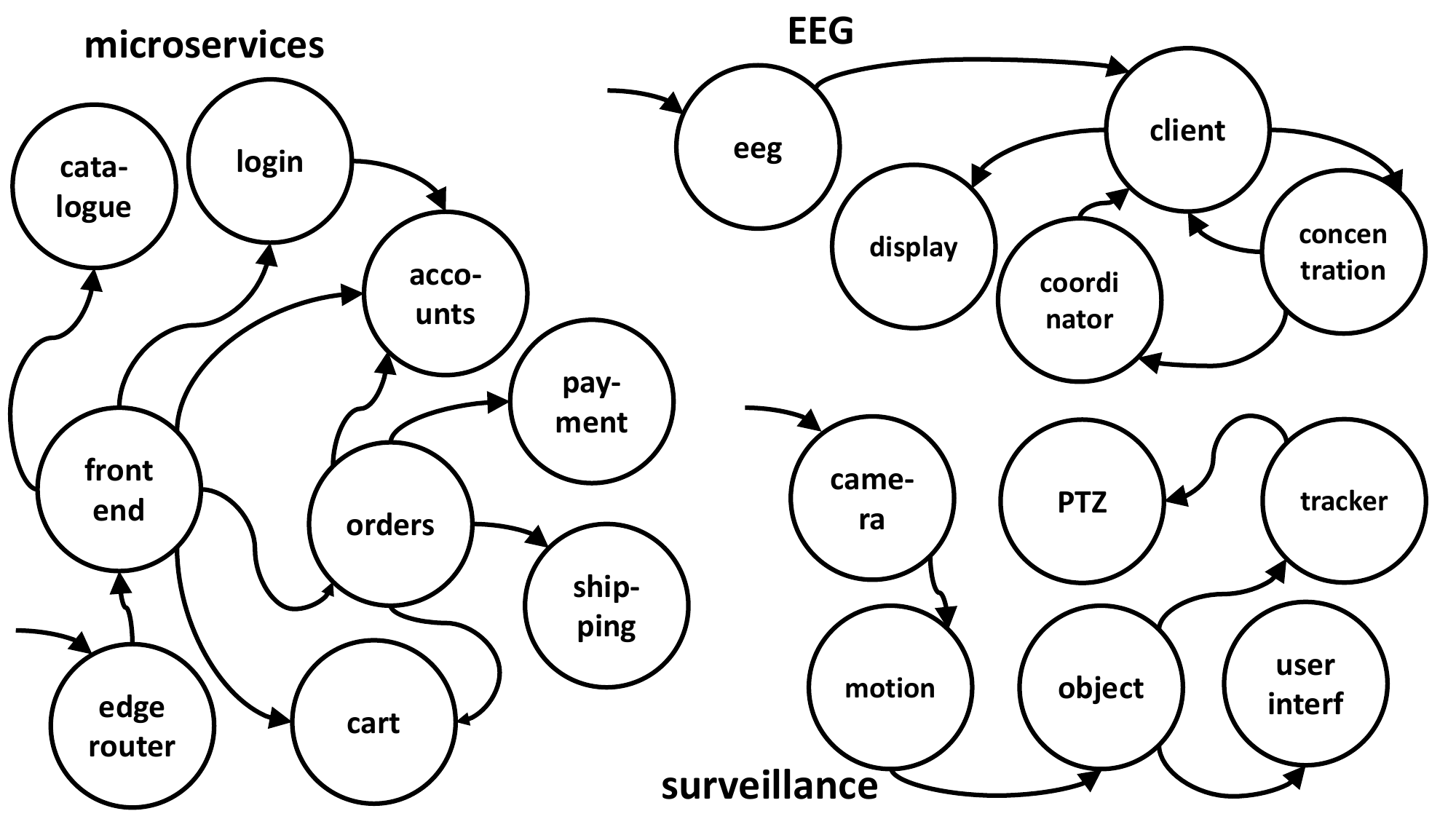}
	\caption{Services interoperability of the three applications.}
	\label{appdefinition}
\end{figure}

Two experiment sizes were considered by changing the number of applications in the system 15, and 30 applications (100, and 200 services respectively). The applications were based on three different types or application templates: a latency-sensitive online EEG (electroencephalography) tractor beam game defined by Zao et al.~\cite{6910482} and used in the experiments of the fog resource policy of Gupta et al.~\cite{SPE:SPE2509}; an intelligent surveillance through distributed camera networks defined by Hong et al.~\cite{Hong:2013:MFP:2491266.2491270} and also used in the experiments of Gupta et al.~\cite{SPE:SPE2509}; and, finally, an e-commerce web application based on microservices~\cite{weaveworks2016shocksshop} that we previously used in several container orchestration studies~\cite{Guerrero2018a,Guerrero2018c}. The services and their interoperability for the three applications can be observed in Figure~\ref{appdefinition}.

The capacity of the fog devices ($R^{cap}_{f_i}$) was uniformly defined in the range of 4-10 resource units. The resource consumption of the services ($R^{con}_{s_x}$) was also defined uniformly, but with values between 1 and 4 resource units. Thus, the maximum number of services in a device is 10 and the minimum is 1.

The network links were characterized with a communication latency $\mathit{L}_{C_{i,i'}}$ between 75 and 125 ms, except for the links with the cloud provider that were fixed in $\mathit{L}_{C_{i,cloud}}=100.0$ ms.

The number of IoT gateways was the 20\% of the total number of fog devices, resulting on 20 gateways for the considered network size. We considered eight users connected to each gateway, so the total number of users in the system was also 160.

The IoT devices were distributed in the IoT gateways by considering a case that the requests for the same application were received from the same region of the network. Therefore, a random IoT gateway was randomly and uniformly selected for each application. The IoT devices requesting the same application were placed in this random gateway and its $k$ nearest neighboring gateways, where $k$ is the number of IoT devices per application. 
The neighbors were determined by considering the shortest path distance between devices.

\subsection{Evolutionary algorithms parametrization}
\label{sectevolalgparameters}

Some of the parameters of the algorithm were set in a preliminary exploratory phase. We tested a range of values and we selected the smallest values with the best performances. The final selected parameters are shown in Table~\ref{algparametrization}.

\begin{table}[t]
	\centering
	\caption{Algorithm parametrization for the experiments.}
	\label{algparametrization}
	\begin{tabular}{lllr}		
		\toprule	
		 \textbf{WSGA}  & \textbf{NSGA-II}  
		&  \textbf{MOEA/D} &\textbf{Value}\\
		\textbf{Parameter} &\textbf{Parameter}  &\textbf{Parameter} &\\

			\midrule
populationNum.&populationNum.&$N$& 100\\	
generationNum.&generationNum.&generationNum.& 400\\
mutationProb&mutationProb&mutationProb& 0.25\\
--&--&$T$& 20\\
$\theta_{spread}$, $\theta_{resource}$, $\theta_{latency}$ &--&--& $\frac{1}{3}$\\
$\omega_{spread}$, $\omega_{resource}$ &--&--& 1.0\\
$\omega_{latency}$ &--&--& \multicolumn{1}{l}{max.}\\
&&& \multicolumn{1}{l}{length}\\
&&& \multicolumn{1}{l}{path}\\

		\bottomrule	
	\end{tabular}
\end{table}

The population size was fixed in 100 solutions as greater populations did not obtain better optimizations. The population size corresponds to the number $N$ of weight vectors for the case of MOEA/D. The number of generations was fixed in 400. We selected a high enough value to be able to detect the stabilization of the objective values. We considered a size $T$ for the neighbor solutions of 20 for the MOEA/D. The preliminary results showed that a mutation probability of 0.25 was enough to achieve a wide diversity of search space.

The weighted sum transformation was fixed with our own preferences. We considered the three objectives equally important and, consequently, the three weights were fixed with the same value:

\begin{equation}
\theta_{spread} =   \theta_{resource} =  \theta_{latency} = \frac{1}{3}
\end{equation}

Additionally, service spread and free resources were already normalized because they were unity-based and their values were between 0.0 and 1.0. The scaling factor was only necessary for the network latency objective. In this case, we normalized by scaling the values of this objective to a range of $[0,1]$, with the formula:

\begin{equation}
x' = \frac{x - x_{min}}{x_{max} - x_{min}}
\end{equation}
considering that the maximum network latency $x_{max}$ is the distance between the cloud provider and its most distance device, the worst case for allocating two interoperated services, and the minimum $ x_{min}$ is 0, the case of placing two interoperated services in the same device.

\section{Results and discussion}
\label{resultdiscussion}

The results are presented in such a way that the three algorithms are easily comparable. It is important to remember that the solution for a multi-objective optimization is usually a set of non-dominated solutions, such in the cases of NSGA-II and MOEA/D. Thus, two analysis are considered in our study: (a) The comparison between one solution selected from the resulting set (Section~\ref{sect:onesolution}). The selection of the solution is done with criteria that can be fixed by the system administrator. Under those criteria, the selected solution is considered the \textit{best} from the solution set; (b) The comparison between the whole set of solutions (Section~\ref{sect:paretoresult}).

\subsection{Analysis of one selected solution}
\label{sect:onesolution}

\begin{figure}[t]
	\includegraphics[width=1.0\linewidth]{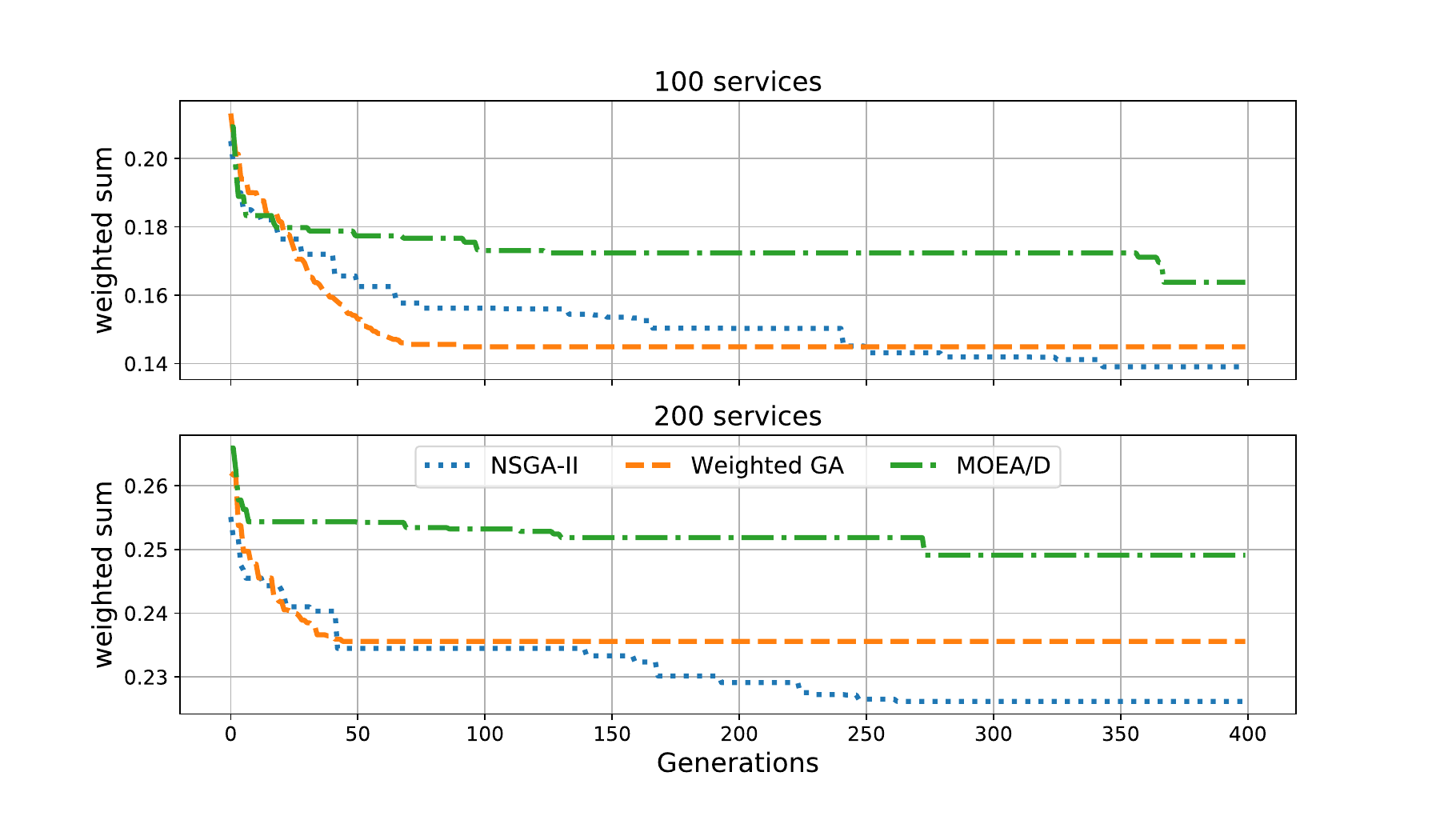}
	\caption{Evolution of the weighted sum of the objectives for the \textit{best} solution for each evolutionary algorithm.}
	\label{totevolution}
\end{figure}

Figure~\ref{totevolution} shows the results along the generations of one solution for each of the three algorithms. This selection was performed under some criteria, and the solution was considered the \textit{best} one from this point of view. Since the WSGA used the uniformly weighted sum transformation for the evaluation of the fitness value (Eq.~\ref{eqgenweighted}), we used these same criteria to select the solution from the Pareto sets obtained with NSGA-II and MOEA/D. Thus, the figure shows the value of the solution with the smallest weighted sum value for each generation. Note that, to improve the visualization of the differences in the results between algorithms in each experiment, the scales of y-axis are different for each plot.

\begin{figure}[t]
	\includegraphics[width=1.0\linewidth]{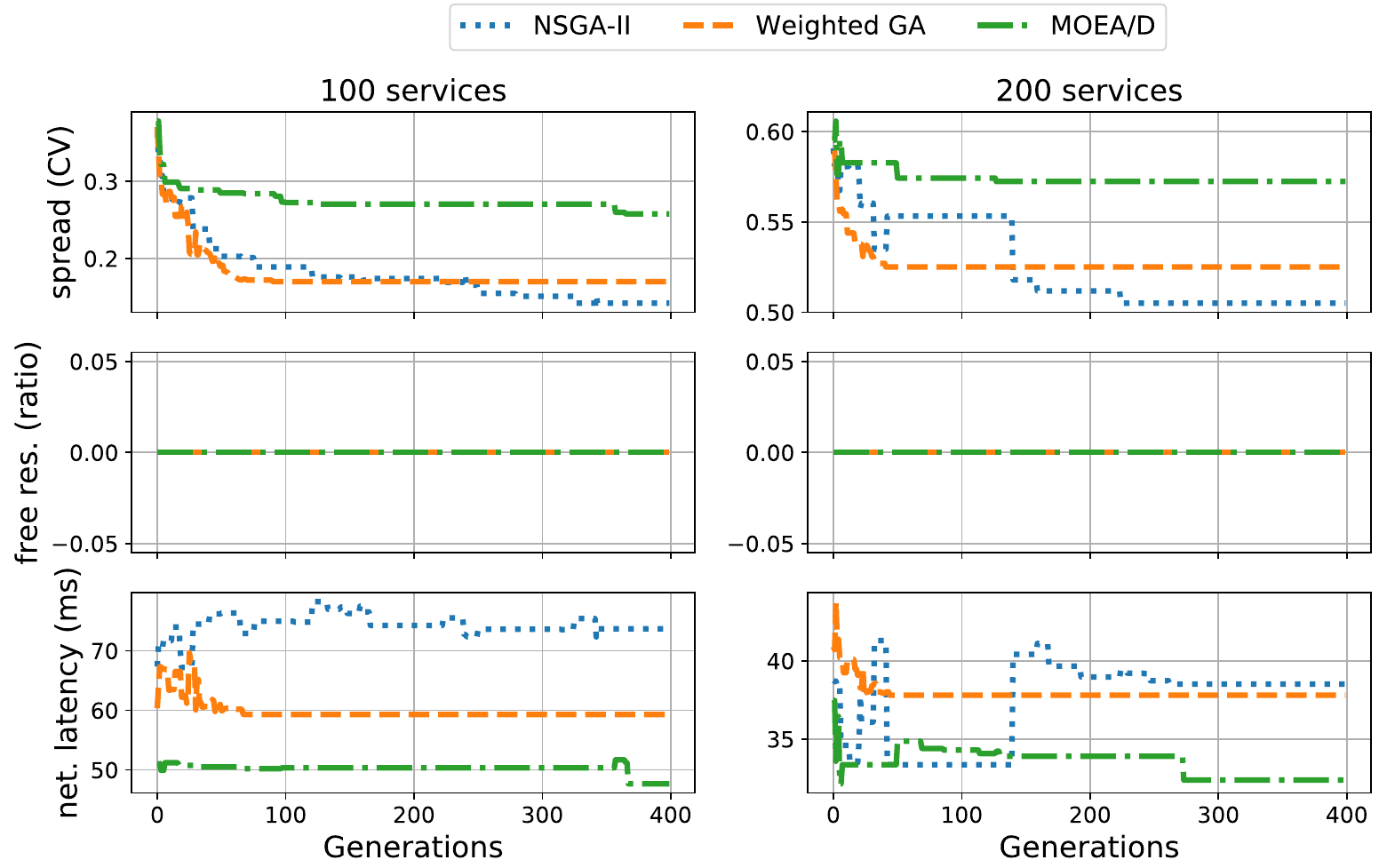}
	\caption{Evolution of the three objectives for the \textit{best} solution for each evolutionary algorithm.}
	\label{objectiveevolution}
\end{figure}

For a deeper comparison of the three selected solutions, we also present Figure~\ref{objectiveevolution}, where the three objectives values are disaggregated. Note that, once again, the scales of the y-axis are different between plots.

From the analysis of the first results, Figure~\ref{totevolution}, the general conclusion is that NSGA-II is the algorithm that obtained the smallest uniformly weighted sum of the objectives values, at least for the \textit{best} selected solution. On the contrary, MOEA/D resulted in the highest weighted sum. Additionally, the NSGA-II algorithm obtained the smallest weighted sum in fewer generations for the case of 200 services (around generation 150) than in the experiment with 100 services (around generation 300).

Considering the results for each single objective (Figure~\ref{objectiveevolution}), we can observe that the algorithms have different behaviors for the three objectives. First, the free resource objective is easily minimized by the three algorithms. It is observed that the \textit{best} solution use all the resources in the system from the very beginning. Second, in the case of the service spread, NSGA-II is the best algorithm. Third, the network latency is more optimized by the MOEA/D. The WSGA is always in a medium-term between the two other algorithms.

It is observed that important optimizations are obtained in the first generations but the values in Figure~\ref{objectiveevolution} are very irregular, mainly for the NSGA-II. Note that evolutionary algorithms are metaheuristics and they are based on the generation of random solutions. The algorithm starts with a random population that is evolved along the generations with random combinations and modification of its solutions. Consequently, at the beginning of the optimization (the first generations), the population is not stabilized, and the solutions are very far from the optimized solution. By this, the changes in the population are very important between generations and, consequently, the values of the optimized solution are irregular. Moreover, if we note that the solutions plotted in the figure are the ones with the smallest weighted sum for each generation, thus the selected solution is not the same one along the generations, mainly in the first ones. On the contrary, when the population is stabilized (last generations), the solutions are closer to the optimized value, and the population is quite similar between generations. Consequently, the selected solution is many times the same solution for different generations because there are not important changes in the population. However, it is important to remember that metaheuristics cannot guarantee to reach the optimized solution. Thus, new optimization could be achieved for higher number of generations. But, these small improvements of the objectives do not justify so large increases in the number of generations.

By the analysis of Figures~\ref{totevolution} and~\ref{objectiveevolution}, generation 300 could be a suitable point to fix the ending condition of the genetic algorithms. However, we can reduce the number of generations if we are interested in reducing the the execution time. For example, with only 50 generations, the most significant improvements are already achieved. But for this reduced number of generations, NSGA-II is not the best solution. 

Finally, if we compare the algorithms between them, NSGA-II seems to be the algorithm that needs more generations to minimize the solutions. On the contrary, MOEA/D is the \textit{fastest} solution.

\subsection{Analysis of the Pareto solution set}
\label{sect:paretoresult}

\begin{figure}[t]
	\includegraphics[width=0.9\linewidth]{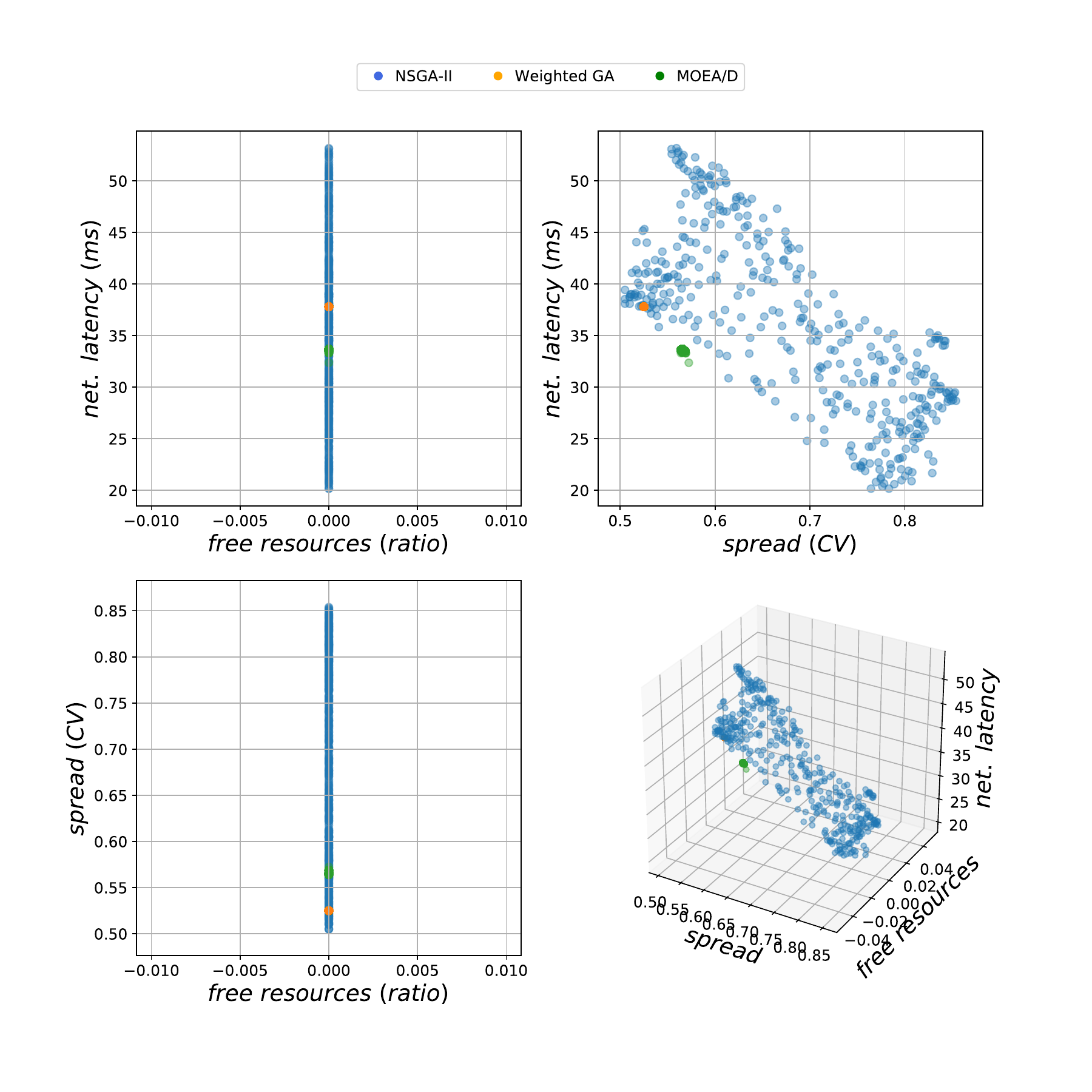}
	\caption{Final set of solutions obtained for the experiment with 200 services.}
	\label{finalpareto1}
\end{figure}

The evaluation of the results obtained with a multi-objective optimization algorithm cannot be only analyzed with one solution from the solution set~\cite{1197687}. Thus, we also include the representation of the final solution sets in Figure~\ref{finalpareto1}. The figure includes the Pareto set of the NSGA-II, the external population of the MOEA/D and the whole population in the case of the WSGA. In the three cases, 100 solutions (points) are represented. We want to recall that a solution corresponds to a placement configuration (the matrix $A$), and the result of the algorithms are a set of optimized solutions. Each point of the 3D plot of the figure represents one of these solutions, characterized by its objective values. The three dimensions of the 3D scatter plot are the three optimization objectives. The other three scatter plots are the 2D projections of the 3D plot for each pair of objectives.

We can observe that the Pareto set of the NSGA-II covers a wider range of solutions, mainly in the case of the network latency and service spread objectives. In the case of the free resources, all the solutions are located in the value of 0.0. This is because this objective is probably the easiest one to be minimized since this minimization is just obtained by placing as many services as possible. On the contrary, the coverage of the solutions of the MOEA/D is very limited to a small region with solutions between 30-35 ms of network latency and 0.55-0.6 of service spread. Finally, all the solutions of the WSGA are located in the same point, i.e., all the solutions have the same objective values. This offers a low flexibility to the system administrator for selecting solutions under different criteria.

The previous analysis of the solution spread is also commonly measured by using the volume of the hyper-cube that envelops the solutions~\cite{wu2000, 1197687}. The hyper-cube is generated by multiplying the wide of the solution space for each single objective, i.e., the distance between the minimum and maximum value for each objective:

\begin{equation}
Solution\ Spread\ Volume = \prod_{i}^{num. objectives} | max(i) - min(i) |
\end{equation}

In our particular case, we need to limit this analysis to the service spread and network latency objectives. If we also consider the resource usage, the solution spread volume will be 0.0 because all the solutions result in values of 0.0 for the free resources. Consequently, we measure the coverage of the solution as $| max(net.\ latency) - min(net.\ latency) | \times | max(spread) - min(spread) |$. This results are reflected in Table~\ref{paretospread2} for both experiment sizes. The values of the table confirms our conclusions from the analysis of the scatter plots.

\begin{table}[t]
	\centering
	\caption{Solution spread volume for the network latency and service spread objectives.}
	\label{paretospread2}
	
	\begin{tabular}{lrrr}		
		\toprule	
		\textbf{Apps}  
		&  NSGA-II &WSGA & MOEA/D \\
		\midrule
100 services &0.1505&0.0&0.0013\\
200 services &0.0519&0.0&0.0001\\

		\bottomrule	
	\end{tabular}
\end{table}

\subsection{Execution times}

\begin{figure}[t]
	\includegraphics[width=1.0\linewidth]{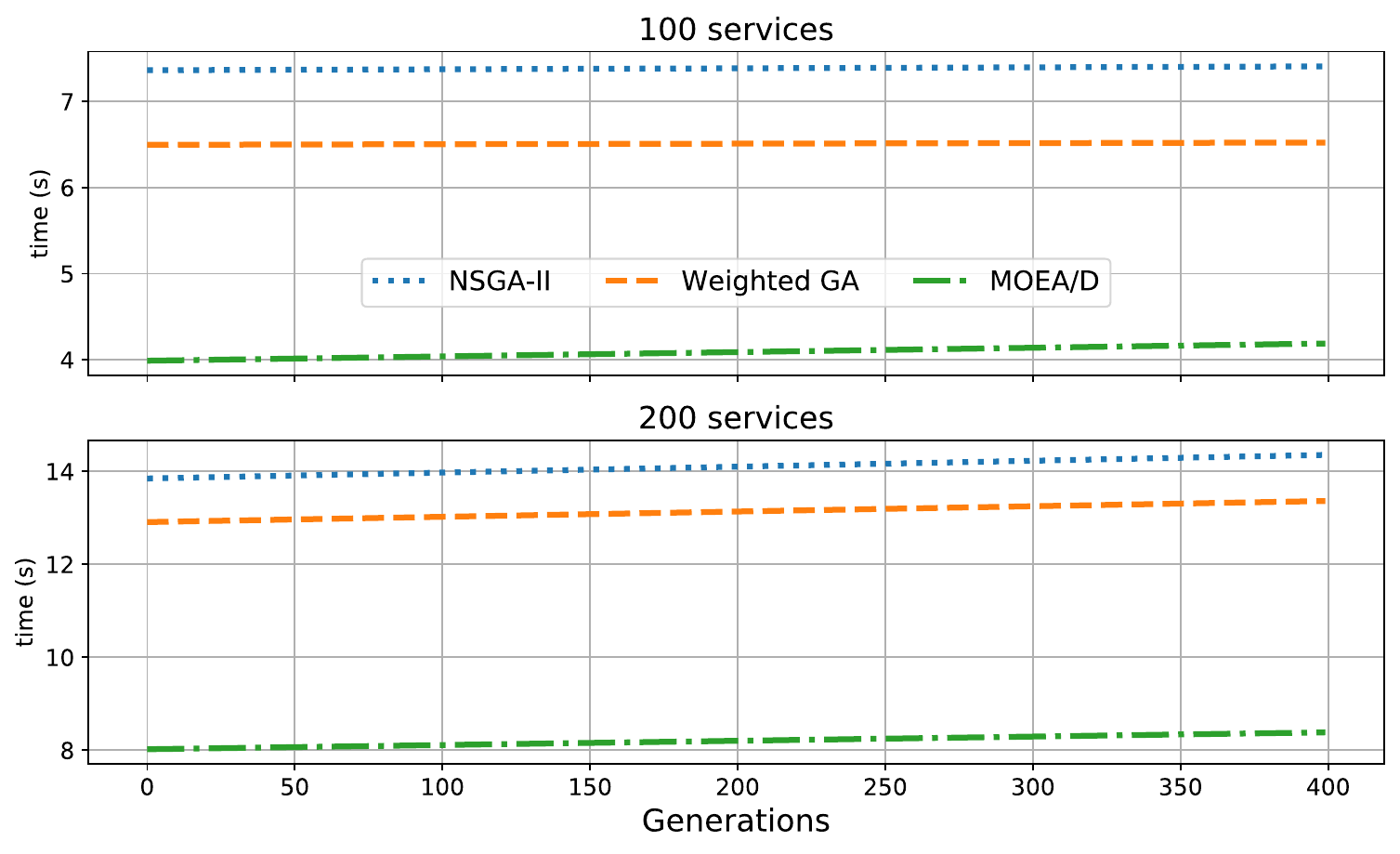}
	\caption{Execution times of the three optimization algorithms.}
	\label{executiontimes}
\end{figure}

Finally, the results of the execution times for the three optimization algorithms are presented in Figure~\ref{executiontimes}.  We implemented the three algorithms in Python 2.7.6. Their source code can be found in a public repository~\cite{thesourcecode}. The experiments were executed on a computer running  MacOS Sierra with an Intel(R) Core(TM) i7 processor operating at 3.10 GHz with 16 GB of RAM.

In general terms, the MOEA/D is much faster than the other two algorithms, and NSGA-II is the one with the highest execution times. Additionally, we can observe that the execution times remain almost constant along the algorithm executions. It is also observed that the execution times of the second experiment (the one with 200 services) are almost doubled with regard to the smallest experiment (100 services).

This increase in the execution time is explained by the differences between these both experiments which are the sizes of the solutions. As we explained in Sections~\ref{systemmodel} and~\ref{geneticoperators}, solutions are represented with a matrix $A$ of size $|S| \times |F|$ where $|S|$ is the number of services and $|F|$ the number of devices. Note that, by the analysis of the source code of the algorithms, the computational complexity of the fitness calculation can be determined as $O(3SF)$. Thus, it is clear that the execution times depend on the experiment sizes, i.e., the number of devices and services. Consequently, the same increase in the execution time is expected for bigger experiments.

\section{Conclusion}
\label{conclusion}

We have evaluated and compared the efficiency of three evolutionary algorithms for the fog service placement problem. The adaptation of the algorithms to our specific domain supposed to define the genetic operators and the parametrization of the algorithms. A model for the fog architecture domain and the definition of applications as a set of interoperated services has been defined. Three objective functions have been formalized for our three main optimization concerns: the increase of the network latency due to the spread of the services across the fog devices (minimization of the network latency), the highest possible use of the fog resources to reduce the use of the cloud resources (minimization of the free resources), and an evenly distribution of the services across the fog devices (optimization of the service spread).

Solutions to FSPP are represented as a matrix that embeds the allocation and scale level of the services to be deployed. The algorithms determine the number of instances of each service and their placement in the architecture in order to minimize the three considered objectives. Consequently, the solutions are defined as a placement plan represented with a matrix of services and devices.

The experimental validation was performed with a random Barabasi-Albert network topology of 100 devices and two experiment sizes of 100 and 200 services. The same patterns in the solution of both cases were observed. In general terms, the NSGA-II algorithm resulted in the algorithm that achieved the highest optimizations (smallest objectives values). But if the objectives are analyzed independently, NSGA-II obtained better results in the service spread and MOEA/D in the network latency.  Moreover, NSGA-II also obtained a wider solution space and, consequently, the system administrator has a higher flexibility to select one solution from the solution set by considering different criteria or preferences. The benefits of the NSGA-II were obtained at the cost of longer execution times, both in terms of the number of generations and execution times of each generation. To sum up, NSGA-II was better to optimize the objectives and to obtain a more diverse solution space. MOEA/D was better to reduce the execution times. The WSGA algorithm did not show any benefit with regard to the other two algorithms.

Research opportunities for future works emerged from the study of a hybrid optimization algorithm that simultaneously executes the three optimization algorithm and merges the three solution sets. Additionally, other metaheuristics, such as swarm particle optimization, ant colony optimization, or firefly optimization, can be also studied and compared between them. Finally, the applicability of these evolutionary solutions to other fog organizations (such as multi-level fog, federated fog, fog colonies\ldots) is interesting, not only for the placement of the services but also for the simultaneous optimization of the placement and the fog organization.

\section*{Acknowledgements}
Funding: This work was supported by the Spanish Government (Agencia Estatal de Investigaci\'on) and the European Commission (Fondo Europeo de Desarrollo Regional) [grant number TIN2017-88547-P MINECO / AEI / FEDER, UE].

\section*{References}

\bibliographystyle{elsarticle-num} 
\bibliography{sample-bibliography}


%
%
%
\end{document}